\documentclass[letterpaper, 10 pt, conference]{ieeeconf}  

\IEEEoverridecommandlockouts 

\usepackage{amsmath}
\usepackage{dsfont}
\usepackage{tabularx}
\usepackage{graphicx}
\usepackage[T1]{fontenc}
\usepackage[utf8]{inputenc}
\usepackage{babel}
\usepackage{blindtext}
\usepackage[dvipsnames]{xcolor}
\usepackage{xcolor,colortbl}
\usepackage{float}

\definecolor{Gray}{gray}{0.85}
\definecolor{LightCyan}{rgb}{0.88,1,1}
\renewcommand\footnotemark{}

\newcolumntype{a}{>{\columncolor{Gray}}c}
\newcommand{\up}[1]{\textsuperscript{#1}}
\definecolor{mypink2}{RGB}{2, 4, 143}

\begin{document}
\title{\LARGE \bf
Interpretable Goal-Based model for Vehicle Trajectory Prediction in Interactive Scenarios
}

\author{Amina Ghoul\textsuperscript{1} ,
 Itheri Yahiaoui\textsuperscript{2}, Anne Verroust-Blondet\textsuperscript{1}, and Fawzi Nashashibi\textsuperscript{1}}


\thanks{This work was carried out in the SAMBA collaborative project, co-funded by BpiFrance in the framework of the Investissement d’Avenir Program. \\
1: INRIA Paris, France {\tt firstname.lastname@inria.fr}\\
2: CReSTIC, Universit\'e de Reims Champagne-Ardenne, Reims, France {\tt itheri.yahiaoui@univ-reims.fr}}


\maketitle
\thispagestyle{empty}
\pagestyle{empty}
\begin{abstract}
The abilities to understand the social interaction behaviors between a vehicle and its surroundings while predicting its trajectory in an urban environment are critical for road safety in autonomous driving. Social interactions are hard to explain because of their uncertainty. In recent years, neural network-based methods have been widely used for trajectory prediction and have been shown to outperform hand-crafted methods. However, these methods suffer from their lack of interpretability. In order to overcome this limitation, we combine the interpretability of a discrete choice model with the high accuracy of a neural network-based model for the task of vehicle trajectory prediction in an interactive environment. We implement and evaluate our model using the INTERACTION dataset and demonstrate the effectiveness of our proposed architecture to explain its predictions without compromising the accuracy.
\end{abstract}
\section{Introduction}

Predicting the future motion of a dynamic agent in an interactive environment is crucial many fields and especially in autonomous driving.  
However, this task is challenging as it depends on various factors such as the agent’s intention or the interaction with his surroundings. Because of these uncertainties, future motion of agents are inherently multimodal.
To ensure safe predictions, the agent needs to take into account the dynamics of the surroundings and timely predict their motions in near future to avoid collisions.
To address the task of forecasting vehicle motion, many studies use neural network-based model. One major drawback of these methods is the lack of interpretability. In fact, although data-driven approaches achieve outstanding performance in various tasks, it is hard to trust and interpret their predictions. For this reason, in recent years, developing models that can understand social interactions and forecast future trajectories has been an active and challenging area of research.

Early works designed hand-crafted methods based upon
domain knowledge to forecast dynamic agents trajectories, either
with physics-based models such as Social Forces \cite{helbing1995social}, or
with pattern-based models such as discrete choice modelling (DCM) \cite{antonini2005discrete}. These models, based on domain
knowledge allow their predictions to be interpretable.

The nature of vehicle movement is highly connected to the motion of other road users around them. They alter their paths  according to their interactions with neighbors. Thus, the concept of social interaction has been highly evaluated and discussed in the existed studies \cite{alahi2016social}. 
Our interest in this problem stems from the fact that while interaction modeling has been well-investigated in
existing studies, it’s hard to interpret the learned social interactions. In these previous studies, variables in models are designed to learn latent behavioral
characteristics and with no expectation to have practical implications. For example, the pooling methods \cite{alahi2016social} directly aggregate hidden states of all neighbors in a neighborhood to learn the connections between people. Thus, it’s hard to understand what kind of social interactions is going on, how it varies among moving pedestrians and how it affects
the future trajectories. Attention mechanism \cite{messaoud2020attention} can show the interests of agents in each neighbor by observing
the learned distribution of attention, thus we know which agent have the greatest influence on agent. However, we still
can’t get a concrete pattern of the social interaction. Therefore, these neural network-based models suffer from the lack of interpretability regarding the model’s decision-making process.

To address these limitations, we propose to combine an interpretable discrete choice model with a neural network for the task of vehicle trajectory prediction.
Our approach presents a way to easily validate NN
models in safety critical applications, by using the
interpretable pattern-based rules from the DCM.
We conduct extensive experimentations on the real-world INTERACTION dataset and we demonstrate the effectiveness of our method, while at the same time providing a rationale behind high-level decisions, an essential component required for safety-critical applications like autonomous systems.
We also conduct a comparative study between two discrete choice models.

\section{Related Work}
\subsection{Knowledge-based Models}
Early works address trajectory prediction problem using of knowledge-based
methods. \cite{treiber2000congested} use Kalman filter to predict vehicle future trajectory. Discrete choice modelling (DCM) uses a grid for selecting the next action relative to each individual. DCMs have been used to predict pedestrian's trajectories  \cite{robin2009specification}, and also for many applications in various fields such as facial expression recognition \cite{antonini2006discrete}.
These knowledge-based methods allow interpretable outputs, but they usually fail to capture the complexity of agent-agent interactions and agent-scene interactions. Therefore, they have low prediction accuracy when predicting trajectories.

\subsection{Data-driven Models}
In order to solve the low accuracy problem of knowledge-based models, in recent years, many studies tackle the task of motion prediction using neural network models \cite{messaoud2020trajectory, zhao2020tnt}. 
\cite{alahi2016social} introduced the social LSTM for pedestrian trajectory prediction. They encode the motion of each agent using an LSTM. Then, they extract
the interactions between agents by sharing the hidden states between all the LSTMs
corresponding to a set of neighboring pedestrians. MHA JAM \cite{messaoud2020trajectory} applies multi-head attention by considering a joint representation of the static scene and surrounding agents. The authors use each attention head to generate a distinct future trajectory to address multimodality of future trajectories.
However, these data-driven methods lack the ability to output predictions that can be explained.

\subsection{Interpretable Trajectory Prediction}
To adress the lack of interpretability in neural network-based models, recent studies focus on adding expert knowledge to deep learning models for trajectory prediction. Neumeier et al. \cite{neumeier2021variational} use an autoencoder where the decoder contains expert knowledge
to produce an interpretable latent space in a vehicle trajectory prediction model, in a highway environment. Another way to encourage interpretability in trajectory prediction architectures is through discrete modes.
For example, Brewitt et al. \cite{brewitt2021grit} propose a Goal Recognition
method by Interpretable Trees (GRIT) where the “goal” is defined as many kinds of behavioral intentions, such as “straight-on”, “turn left”, “u-turn”, and “stop”, etc.
This aims the goal recognition to be interpretable by humans.
Kothari et al. \cite{kothari2021interpretable} learn a probability distribution over possibilities in an interpretable discrete choice model for the task of pedestrian trajectory prediction.
 
We use a similar approach for the task of vehicle trajectory prediction in an urban environment. However, unlike \cite{kothari2021interpretable}, we first predict the goal and then the whole trajectory for a prediction horizon greater than 1 second.

To the best of our knowledge, we are the first to use a DCM to help model the behavior of vehicles in their interactions with their surroundings.

In this paper we also consider and compare two types of discrete choice models describing the behavior of vehicles.

\section{Method}
\label{model}
\subsection{Problem definition}
The goal is to predict the future trajectories of a target agent $T$ : $\hat{Y_T}=(\hat{x}_T^t, \hat{y}_T^t)$ from time $t = t_{obs}+1$ to $t = t_f$.
We have as input of our model the track history of the target agent and the $n$ neighboring agents in a scene defined as 
$\textbf{X} = [X_1, X_2, ..., X_n]$. Each agent $i$ is represented by a sequence of its states, from time $t=1$ to $t=t_{obs}$. Each state is composed of a sequence of the agent relative coordinates $x_i^t$ and $y_i^t$, velocity $v_i^t$, acceleration $a_i^t$, heading $\theta_i^t$.
\begin{eqnarray}
X_i^t = (x_i^t, y_i^t, v_i^t,a_i^t, \theta_i^t)
\end{eqnarray}
The positions of each agent $i$ are expressed in a frame where the origin is the position of the target agent at $t_{obs}$. The y-axis is oriented toward the target agent’s direction of motion and x-axis points to the direction perpendicular to it.

\subsection{Discrete Choice Model}

Discrete choice models or DCMs are hand-crafted models used to explain or predict a choice from a set of alternatives $K$ made by a decision-maker.
DCMs are knowledge based models that have a high interpretability. However, despite having interpretable outputs, these models suffer from low prediction accuracy. For that reason, \cite{kothari2021interpretable} proposed a model combining the high interpretability of the DCMs and the high accuracy of the neural network-based model to predict pedestrian's trajectories. In this paper, we present an architecture that can model the interactions between vehicles and their surroundings. \\
We use the Random Utility Maximization (RUM) theory \cite{manski1977structure} that postulates that the decision-maker aims at maximizing the utility relative to their choice. The utility that an agent $i$ chooses an alternative $k$, is given as :
\begin{eqnarray}
U_{ik} = \sum_d \beta_d b_{dik} + \epsilon_{ik},
\end{eqnarray}
where $\beta$ are the parameters associated with the explanatory variables $b$ that describe the observed attributes of the choice alternative. We assume that the random terms $\epsilon_{ik}$ are independently and identically distributed (i.i.d.) follow an Extreme Value Type I distribution with location parameter zero and the scale parameter 1.  
In our case, we propose and compare two utility functions $u_k$ for an alternative $k$. These functions are defined and explained in details in Section \ref{dcms}.
The alternative $k$ corresponds to the target agent’s goal at timestep $t_f$, extracted from a radial grid, similar to \cite{kothari2021interpretable}.

\subsection{Neural Networks Model}

\begin{figure*}[ht]
\centering
\includegraphics[scale=0.40]{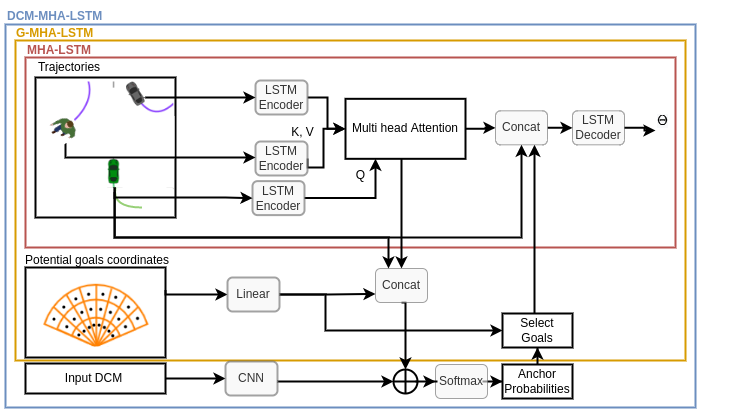}
\caption{Architecture of the compared methods for trajectory prediction. The models take as inputs the past trajectories of the agents in the scene (MHA-LSTM), the target coordinates sampled from a radial grid (G-MHA-LSTM), as well as the input of the DCM model (DCM-MHA-LSTM). They output $L$ trajectories. For more details see section \ref{model}.}
\label{nn}
\vspace{-6mm}
\end{figure*}

For a target agent $T$ at time $t$, $X_T^t$ is embedded using a fully connected layer to a vector $e_i^t$ and encoded using an LSTM encoder,
\begin{equation}
h_i^t = LSTM(h_i^{t-1}, e_i^t; W_{enc}),
\end{equation}
$W_{enc}$ are the weights to be learned. The weights are shared between all agents in the scene.

Then we build a social tensor similar to \cite{messaoud2020trajectory}. We define the interaction space of a target vehicle $T$ as the area centered on its position at $t_{obs}$ and oriented in its direction of motion. We divide this interaction space into a spatial grid of size $(M, N)$. The trajectory encoder states of the surrounding agents $h_i^{t_{obs}}$ are placed at their corresponding positions in the 2D spatial grid, giving us a
tensor $F_s$ of size $(M, N, C_h)$, where $C_h$ is the size of the
trajectory encoder state. 

We use the multi-head attention mechanism \cite{vaswani2017attention} to model the social interactions, where the target vehicle $h_T^{t_{obs}}$ is processed by a fully connected layer to give the query and the social tensor is processed by $1 \times 1$ convolutional layer to give the keys and the values. 

We consider $K$ attention heads where $K$ attention heads are specialized to the $K$ potential goals. 

For each attention head, we concatenate the output of the multi-head attention module $A_k$ with the target vehicle trajectory encoder state $h_T^{t_{obs}}$
to give a context representation $z_k$ for $k=1,... K$.
\begin{equation}
    z_k= Concat(h_T^{t_{obs}}, A_k)
\end{equation}

In order to help the knowledge-based model DCM capture the long term dependencies and the complex interactions, we use the Learning Multinomial Logit (L-MNL) \cite{sifringer2020enhancing} framework.\\
The goal selection probabilities is defined as :
\begin{equation}
\pi(a_k|\textbf{X}) = \frac{e^{s_k(\textbf{X})}}{\sum_{j \in K}e^{s_j(\textbf{X})}},
\end{equation}
where 
\begin{equation}
s_k(\textbf{X}) = u_k(\textbf{X}) + z_k(\textbf{X}),
\end{equation}

where $s_k(\textbf{X})$ represents the goal function containing the NN encoded terms, $z_k(\textbf{X})$, as well as utility function $u_k(\textbf{X})$, following the L-MNL framework.

We consider $L$ attention heads, for each attention head, we concatenate the output of the multi-head attention module $A_l$ with the target vehicle trajectory encoder state $h_T^{t_{obs}}$
to give a context representation $c_l$ for $l=1,... L$.
\begin{equation}
    c_l= Concat(h_T^{t_{obs}}, A_l)
\end{equation}
We select the $L$ best scored targets, and we concatenate their embedding to the output of the context representation $c_l$ for $l =1,... L$. 

Finally, the context vector $c_l$ is fed to an LSTM Decoder which generates the predicted parameters of the distributions over the target vehicle’s estimated future positions of each possible trajectory for next $t_f$ time steps,
\begin{equation}
\Theta_l^t = \Lambda(LSTM(h_l^{t-1}, z_l; W_{dec})),
\end{equation}
where $W_{dec}$ are the weights to be learned, and $\Lambda$ is a fully connected layer.
Similar to \cite{messaoud2020trajectory}, we also output the probability $P_l$ associated with each mixture component.

\subsection{Loss function}

Our proposed model (DCM-MHA-LSTM) outputs the means and variances $\Theta_l^t = (\mu_l^t, \Sigma_l^t)$ of the Gaussian distributions for each mixture component at each time step. \\
The loss for training the model is composed of a regression loss $L_{reg}$ and two classification losses $L_{score}$ and $L_{cls}$. \\
$L_{reg}$ is the negative log-likelihood (NLL) similar to the one used in \cite{messaoud2020trajectory} and given by : 
\begin{equation}
L_{reg} =  -\underset{l}{min}\sum\limits_{t=t_{obs} +1}^{t_{obs}+t_f}  log(\mathcal{N}(y^t|\mu_l^t; \Sigma_l^t))). 
\end{equation}
$L_{score}$ is a cross entropy loss defined as :
\begin{equation}
    L_{score} = -\sum_{l=1}^{L}\delta_{l*}(l)log(P_l), 
\end{equation}

where $\delta$ is a function equal to 1 if $l = l*$ and 0 otherwise.

$L_{cls}$ is also a cross entropy loss defined as :
\begin{equation}
    L_{cls} = -\sum_{k=1}^{K}\delta_{k*}(k)log(p_k), 
\end{equation}
where $p_k$ is the probability associated with the potential goal $k$, $\delta$ is a function equal to 1 if $k = k*$ and 0 otherwise, $k_*^t$ is the index of the potential goal most closely matching the endpoint of the ground truth trajectory. 

Finally, the loss is given by :
\begin{equation}
L = L_{cls} + L_{reg} + L_{score},
\end{equation}
\vspace{-9mm}
\begin{figure}
\end{figure}
\section{Experiments}
\subsection{Dataset}
We evaluate our model on the INTERACTION \cite{zhan2019interaction} dataset.
The INTERACTION dataset
provides a large set of challenging intersection, roundabout,
and highway merge scenarios.  In total, the data is collected from 11 locations using drones or fixed cameras.

\subsection{Compared Methods}
\label{methods}
The experiment includes a comparison of different models:
\begin{itemize}
\item \textbf{I) MHA-LSTM \cite{messaoud2020attention}:} This model only takes as inputs the past trajectories of the agents in the scene and outputs $L$ trajectories with their associated probabilities (see the architecture in the red rectangle in Fig. \ref{nn}). We use $L=6$ attention heads.
\item \textbf{II) G-MHA-LSTM \cite{ghoul2022lightweight}:}
We add to the previous model a radial grid representation from which we extract potential goals. We predict the goal and then the trajectories conditioned on the predicted goal. (see the architecture in the orange rectangle in Fig. \ref{nn}).
\item \textbf{III) DCM-MHA-LSTM :}
To predict the goal of the target agent, we combine the DCM and the neural network using the LMNL framework \cite{sifringer2020enhancing}. 
 This model is described in Section \ref{model} and the architecture is illustrated in the blue rectangle in Fig. \ref{nn}.
\item \textbf{IV) ODCM-MHA-LSTM :} 
This model only uses the DCM to predict the goal of the target agent.
\end{itemize}
\par{\textbf{Goal set representations :}} 
We also compare different types of radial grids.
For the methods II), III) and IV), we compare our results for two types of radial grid : a \textbf{dynamic} grid (d) and a \textbf{fixed} one (f).
Similar to \cite{kothari2021interpretable}, we build the dynamic grid by considering the target agent's current velocity $v_T^{t_{obs}}$.  If $v_T^{t_{obs}}=0$, we replace it with an arbitrary value equals to $0.5$ $m.s^{-1}$.
The fixed grid is built using the value $v= 5.83 m.s^{-1}$, which corresponds to the mean of the velocities in the INTERACTION training set.

\subsection{Compared DCMs}
\label{dcms}
We compare two types of DCMs for modelling the behavior of vehicle motion. For our case, the functions modelling vehicle motion
phenomenon which we consider for goal selection in this
work are:
\begin{enumerate}
    \item \textit{occupancy:} directions containing neighbours in the vicinity are less desirable.
\item \textit{keep direction:} vehicles tend to maintain the same
direction of motion.
\item \textit{collision avoidance:} when a neighbour vehicle’s trajectory is head-on towards a potential goal, this goal becomes less desirable due to the chance of a collision.
\end{enumerate}

\begin{itemize}

\item \textbf{1) DCM 1 : }
\label{dcm1}
For the first DCM configuration, we use a utility function defined as: 
\begin{align}
u_k(\textbf{X}) & = \beta_{dir}dir_k + \beta_{occ}occ_k + \beta_{col}col_k
\end{align}
\label{uk1}
Where the functions $dir_k$, $occ_k$, and $col_k$ correspond respectively to \textit{keep direction}, \textit{occupancy} and \textit{collision avoidance}. These functions are defined in \cite{antonini2005discrete} and \cite{robin2009specification}.
\item \textbf{2) DCM 2 : }
\label{dcm2}
For the second DCM, the utility function is defined as :

\begin{align}
u_k(\textbf{X}) & = \beta_{dir}dir_k + \beta_{occup}occup_k 
\end{align}
\label{uk2}
Where the function $dir_k$ is the same as in \eqref{uk1}. For $occup_k$, we use the same mathematical formula as the occupancy function in \eqref{uk1}, however, we don't consider the position of the neighbors at time $t_{obs}$. Instead, we consider their predicted position at time $t_{obs}+t_f$ using a Constant velocity model. We assume that before predicting his goal, the target agent first predicts the future positions of his surroundings according to their headings and current velocitites, and then avoids the zones that are expected to be crowded. While training this model, we calculate the $occup_k$ function using the grouth truth positions of the neighbors.
\end{itemize}
\subsection{Implementation details}
We use $K = 15$ number of potential goals. Similar to \cite{messaoud2020trajectory}, our interaction space is 40 m ahead of the target vehicle, 10 m behind and 25 m
on each side. We consider the neighbors situated in the interaction space at $t_{obs}$. We also take into account the neighbors that are susceptible of being in this space from time $t_{obs}$ to $t_f$. To do so, we predict the trajectories of all of the neighbors in the scene using a Constant Velocity model and if they have a predicted position in the interaction space, we consider them in our model. We argue that this representation allows to consider neighbors that are not situated in the grid at $t_{obs}$ but that can appear in the grid from time $t = t_{obs}+1$ to $t = t_f$. without having to create a bigger interaction space which can be more computationally expensive. We use $L+K = 6+15$ parallel attention operations. We use a batch size of 64 and Adam optimizer. The model is implemented using PyTorch \cite{paszke2019pytorch}.
\section{Results}
\subsection{Evaluation metrics}
Our method for trajectory forecasting is evaluated with the following three error metrics:
\begin{itemize}
    \item \textbf{Minimum Average Displacement Error over k ($minADE_k$)} : The average of pointwise L2 distances between the predicted trajectory and ground truth over the k most likely predictions.
    \item \textbf{Minimum Final Displacement Error over k ($minFDE_k$)} : The final displacement error (FDE) is the L2 distance between the final points of the prediction and ground truth. We take the minimum FDE over the k most likely predictions and average over all agents.
    \item \textbf{Collision II - Groundtruth collision (Col-II)} \cite{kothari2021human}: This
metric calculates the percentage of collision between the
primary vehicle’s prediction and the neighbors in the
groundtruth future scene.
    
\end{itemize}
\begin{figure*}[ht]
\centering
\includegraphics[scale=0.4]{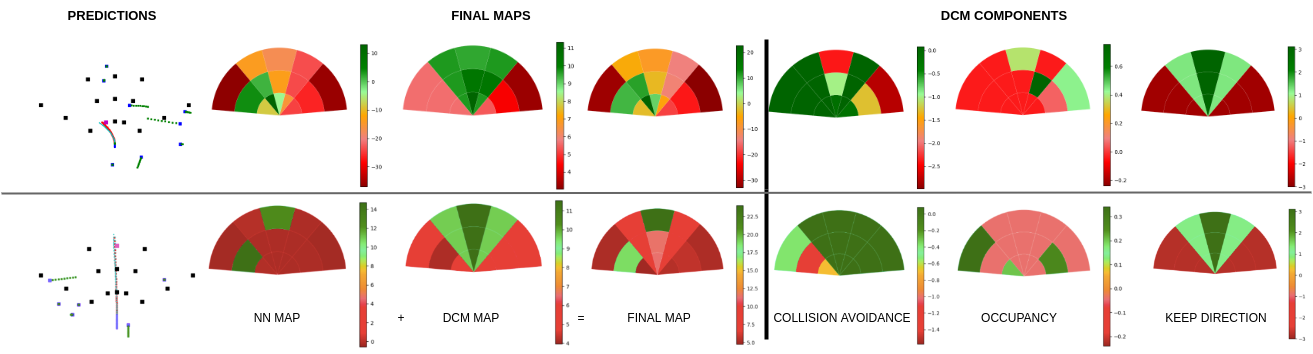}
\caption{Qualitative illustration of the ability of our architecture to output high-level interpretable goals. The potential goals are shown in black and the predicted goal in shown in magenta. The ground truth trajectory is in red and the predicted trajectory is in cyan. Current neighbour positions are shown in blue and their past trajectories are shown in green. In the first row, the decision of the model is influenced by the neural-network (NN). In the second row,
the decision of the model is strongly influenced by the keep direction map of the DCM. (Un)favourable potential goals are shown in green (red).}
\label{dmcviz}
\vspace{-4mm}
\end{figure*}
\subsection{Comparison of Methods}
We compare the methods described in Section \ref{methods}. 

The results are reported in Table \ref{comp_met}. DCM\up{1} and DCM\up{2} refers to the first (resp the second) type of DCM described in \ref{dcms}. (f) and (d) correspond to respectively, the fixed and the dynamic radial grid representation for the extraction of potential goals. We can see that adding the DCM module decrease the percentage of collisions. We can see that the models using a fixed grid perform slightly better than when using a dynamic one. Thus, we can conclude that adding the information about the velocity doesn't improve the results. For future work, we can try multiple grid configurations, which could potentially improve the results.
We can see that using the DCM alone gives worst results as this is due to the fact that the DCM without the NN is not able to predict accurately the goal of the target agent, and therefore, is not able to predict accurate trajectories. We can see that when using the second type of DCM \ref{dcms}, the results of ADE/FDE are similar to the ones using the first type of DCM, however, we notice that the percentage of collisions is lower, indicating that in this case, the utility function is more appropriate to avoid collisions.

\subsection{Comparison with the state-of-the-art}
We compare our approach with the state-of-the-art using the INTERACTION dataset. Our proposed model does not include any map information. In fact, our aim in this paper, is to study the social interactions between the target agent and his surroundings. Therefore, in Table \ref{soa}, are reported the results where we compare our approach with methods that do not use any map information as well such as DESIRE \cite{lee2017desire} and Multipath \cite{chai2019multipath}. We can see that our method outperforms these two models. We then compare our approach against state-of-the-art methods that use map information SAN \cite{janjovs2022san}, TNT \cite{zhao2020tnt}, ITRA \cite{scibior2021imagining} and ReCoG \cite{mo2020recog}. The results are reported in Table.
\ref{soa1}. The main scope is not to compare
the approach to the currently best performing trajectory
prediction networks. The scope
here is to introduce a discrete choice model that provides interpretability, show its feasibility and evaluate
the potential of its prediction performance. Nonetheless, our method still achieves competitive results against these methods. Our model is able to perform well while, unlike any of these methods, providing interpretability.
\begin{table}[ht]
\begin{center}
\caption{Comparison of different methods on the INTERACTION validation set (3 secs horizon) }
\begin{tabular}{c c c c }
\hline
\rowcolor{Gray}
Model & \textbf{$MinADE_6$} &  \textbf{$MinFDE_6$} & \textbf{$Coll-II$} \\
\hline
MHA-LSTM & 0.23 & 0.69 & 6.1 \%\\ 
\hline
G-MHA-LSTM (f) & 0.21 & 0.58 & 1.5 \%\\ 
G-MHA-LSTM (d) & 0.22 & 0.63 & 1.9 \%\\ 
\hline
ODCM\up{1}-MHA-LSTM (f) & 0.20 & 0.58 & 1.4 \% \\
ODCM\up{1}-MHA-LSTM (d) & 0.24 & 0.68 & \textbf{1.0} \% \\
ODCM\up{2}-MHA-LSTM (f) & 0.24 & 0.67 & 1.2 \%\\
ODCM\up{2}-MHA-LSTM (d) & 0.31 & 0.81 & 1.3 \%\\
\hline
DCM\up{1}-MHA-LSTM (f) & \textbf{0.19} & \textbf{0.57} & 1.4 \%\\
DCM\up{1}-MHA-LSTM (d) & 0.20 & 0.60 & 1.4 \% \\
DCM\up{2}-MHA-LSTM (f) & 0.21 & 0.57 &  1.2\%\\
DCM\up{2}-MHA-LSTM (d) & 0.22 & 0.61 & 1.1 \% \\
\hline
\label{comp_met}
\end{tabular} 
\end{center}
\vspace{-9mm}
\end{table}
\begin{table}[ht]
\begin{center}
\caption{Comparison with approaches that do not use the map. }
\begin{tabular}{c c c }
\hline
 & \scriptsize{$MinADE_6$} &  \scriptsize{$MinFDE_6$} \\
\hline

DESIRE \cite{lee2017desire} & 0.32 & 0.88  \\
Multipath \cite{chai2019multipath} & 0.30 & 0.99  \\
Ours & \textbf{0.19} & \textbf{0.58} \\
\hline
\label{soa}
\end{tabular} 
\end{center}
\label{Tab:soa}
\vspace{-6mm}
\begin{center}
\caption{Comparison with approaches that use the map.}
\begin{tabular}{c c c }
\hline
 & \scriptsize{$MinADE_6$} &  \scriptsize{$MinFDE_6$} \\
\hline
SAN \cite{janjovs2022san} & \textbf{0.10} & \textbf{0.29} \\
TNT \cite{zhao2020tnt} & 0.21 & 0.67 \\ 
ITRA \cite{scibior2021imagining} & 0.17 & 0.49 \\
ReCoG \cite{mo2020recog} & 0.19 & 0.66 \\
Ours & 0.19 & 0.58  \\
\hline
\label{soa1}
\end{tabular}
\end{center}
\label{Tab:soa1}
\vspace{-10mm}
\end{table}

\subsection{Interpretable outputs}
\subsubsection{Estimation of $\beta$}
We  study the coefficients $\beta$ of the utility function of our DCM obtained by training our model.
\begin{table}[ht]
\begin{center}
\caption{Estimation of $\beta$}
\begin{tabular}{c c c c c}
\hline
\rowcolor{Gray}
Model & \textbf{$\beta_{dir}$} &  \textbf{$\beta_{col}$} & \textbf{$\beta_{occ}$} & \textbf{$\beta_{occup}$} \\
ODCM\up{1}-MHA-LSTM (f) & -6.7 & -0.7 & 1.2  & - \\
ODCM\up{2}-MHA-LSTM (f) & -10.4 & - & - &-0.3\\
\hline
DCM\up{1}-MHA-LSTM (f) & -2.3 & -0.3 & 0.2 & - \\
DCM\up{2}-MHA-LSTM (f) & -2.4 & - & - & -0.1\\
\hline
\label{comp_beta}
\end{tabular} 
\end{center}
\vspace{-10mm}
\end{table}

The estimated parameters of both of the utility functions Eq. \ref{uk1} and Eq.\ref{uk2}  are reported in Table. \ref{comp_beta} for a fixed radial grid.
We can see that the all of the coefficients $\beta_{dir}$ are negative. This means that the utility of an alternative is going to decrease when its angular position is more decentralised with respect to the current direction, respectively. This is coherent as vehicles tend to keep their direction.
The $\beta_{occ}$ parameters have a positive sign, implying that vehicles tend to prefer nearby spatial zones crowded by agents. This result is not coherent as we would expect the contrary. However, this result can be interpreted as in situations where a lot of agents are moving toward the same destination. 
The $\beta_{col}$ parameters have a negative sign. This means that vehicles tend to avoid zones where there are potential colliders.
We can see that the coefficients $\beta_{occup}$ and $\beta_{dir}$ are negatives. This is coherent as vehicles tend to keep their direction and avoid occupied zones.

\subsubsection{Interpretability of the Goals}

We demonstrate the ability of our network to output interpretable goals in Fig. \ref{dmcviz}.  In addition to the predictions map, we illustrate the activation maps of : the neural network (NN) map, the overall DCM map and finally the DCM function for the first DCM configuration described in Section. \ref{dcm1}. In the first row of Fig. \ref{dmcviz}, the NN map influence the most the final decision. However, in the second row the influence of the NN map is weaker and the decision is more influenced by the DCM map. We can see that the decision is mostly influenced by keep direction and collision avoidance. Especially, the collision avoidance map helps counteract the influence of the NN map and avoid the model of making the wrong decision. Thus, we observe
that the DCM maps work well in conjunction with the NN
map to provide interpretable outputs.

\section{Conclusion and future work}
In this paper we proposed an interpretable goal based method for the task of vehicle trajectory prediction. In this approach, the discretized goals are selected using both interpretable knowledge-based functions and neural network predictions from the scene. This method allows to combine the high accuracy of the neural networks while being able to understand which vehicle motion rules are present in predicting its goal. Through experiments on the INTERACTION dataset, we highlight the interpretability as well as the accurate predictions outputted by our model. 
As future work, we plan to add lane information to the radial grid in order to make our model more scene-compliant. 
Moreover, we plan to explore other types of utility functions for the DCM to better model the behaviour of vehicles in an interactive environment.
\vspace{-3mm}
\addtolength{\textheight}{-7cm}   





\bibliographystyle{IEEEtran.bst} 
\bibliography{refs}

\end{document}